# Automatic Ship Classification Utilizing Bag of Deep Features

Sadegh Soleimani Pour
Department of Computer Engineering
Payam Noor University (PNU), North
Tehran Branch
Tehran, Iran
sasol_p@yahoo.com

Ata Jodeiri
School of Electrical and Computer ,
University College of Engineering
Engineering
University of Tehran
Tehran, Iran
ata.jodeiri@ut.ac.ir

Hossein Rashidi
Department of Computer Engineering
Sharif University of Technology
Tehran, Iran
h.rashidi128@gmail.com

Seyed Mostafa Mirhassani
Department of Biomedical Engineering
University of Malaya
Kuala Lumpur, Malaysia
mostafamirhassani@gmail.com

Hoda Kheradfallah
Department of Electrical Engineering
Sharif University of Technology
Tehran, Iran
hoda213kherad@gmail.com

Hadi Seyedarabi
Faculty of Electrical and Computer
Engineering
University of Tabriz
Tabriz, Iran
seyedarabi@tabrizu.ac.ir

*Abstract—* Abstract: Detection and classification of ships based on their silhouette profiles in natural imagery is an important undertaking in computer science. This problem can be viewed from a variety of perspectives, including security, traffic control, and even militarism. Therefore, in each of the aforementioned applications, specific processing is required. In this paper, by applying the "bag of words" (BoW), a new method is presented that its words are the features that are obtained using pre-trained models of deep convolutional networks. , Three VGG models are utilized which provide superior accuracy in identifying objects. The regions of the image that are selected as the initial proposals are derived from a greedy algorithm on the key points generated by the Scale Invariant Feature Transform (SIFT) method. Using the deep features in the BOW method provides a good improvement in the recognition and classification of ships. Eventually, we obtained an accuracy of 91.8% in the classification of the ships which shows the improvement of about 5% compared to previous methods.

## I. Introduction

Detection and classification of ships based on natural imagery information and according to their silhouette-profiles are among the challenges that have been of interest to researchers in the field of object recognition. This subject has several potential applications in a variety of fields ranging from security and control to marine navigation.

Object detection and classification is a common yet complex issue in machine vision and image processing. Object detection is the act of processing as well as finding relevant object structures within an image. For each object in the image, there can be a number of descriptive features (i.e. geometric or abstract features) that can be used to identify the underlying structures. Most commonly, the descriptive features are utilized to train mathematical models for the task of object detection [2] [3]. Descriptive features such as Scale Invariant Feature Transform (SIFT) [4], Speeded Up Robust Features (SURF) [5], Local Binary Patterns (LBP) [6], and Binary Robust Independent Elementary Features (BRIEF) [7] are among the most popular features used in the recent years.

To the best of our knowledge, most of the studies in the field of ship detection are based on SAR (Synthetic Aperture

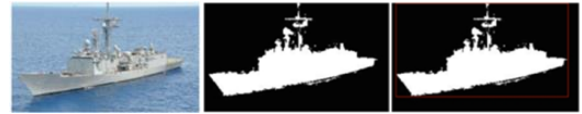

Fig. 1. Step by step results of segmentation algorithm in [17]. From left to right: Raw input image, Image transformed to HSV space and Ship being segmented in a red bounding box.

Radar) imagery [8] [9], [10]. Hu et al. [11] introduced a Cumulative Projection Curve (CPC) approach which can be used to obtain a number of small vessels along the coastline of the ports by defining the overlap degree between the two classes and their method of calculation. Hong et al. [12] presented a hierarchical algorithm for combining the classes that simultaneously performs the diagnosis and segmentation of color images containing the ships. The authors used wavelet-based noise depression techniques, Principal Component Analysis (PCA), and Bayes classification to classify the floating targets. However, some studies criticized the applicability of statistical-based methods like PCA [13]. Other works were performed to detect ships in the wavelet domain with the focus on SAR images [14] [15]. Therefore, they are not applicable to RGB camera images.

The first study for ship detection based on the silhouette-profiles was presented back in 1997 by Gouaillier et al. [16]. In this method, first, PCA is applied to the images. Then a subset of eigenvectors is selected as the feature space that corresponds to the largest eigenvalues. In the recognition process, each input image is mapped to this space, and then, a similarity measurement criterion is used to identify the closest class by assuming the center vectors of each class. Selvi et al. [17] proposed a technique based on the use of form and texture features. Among the shape-related features utilized in this approach, one can name convexness, compactness [18] and Hu's momentum. Multi-scale Gaussian Differential Features, wavelet transformation features, and statistical properties (e.g. variance and entropy) are the texture-related features used in this study. In this work, the image was transferred into the Hue Saturation Value (HSV) space and the segmentation was performed on the image based on the sea color and its boundary with the ship (Fig. 1).

As the next step in [17], the segmented image was down-sampled to a certain size. The important point in their study

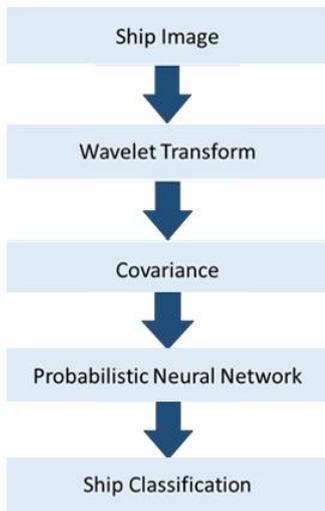

Fig. 2. Structure of the algorithm in [20]. The first image is transformed into the wavelet domain, and then the covariances of wavelet matrixes are calculated and are given as input to the neural network. The maximum output of the network shows the class of the input query.

was the simultaneous use of a real data set and a synthetic dataset prepared through 3D modeling.

Fallahi et al. [20] proposed a method based on the use of an artificial neural network that received the covariance matrix of the wavelet transform of the image as input. The structure of this algorithm is shown in Fig. 2.

Hariri et al. [1] presented a method based on the use of SIFT features and the Bag of Words (BOW) model. Regarding that SIFT features show relatively good robustness to resizing and noise, they used the BoW method to represent these features. Finally, the Support Vector Machine (SVM) was used for classification. They considered five classes of ships in their dataset, all of which were collected from the internet search engines.

Dao-Duc et al. [21] presented an approach using a deep learning architecture like Alex-Net [22]. The network was trained over 130,000 natural images of ships in 35 different categories. The reported accuracies for Top1 and Top5 were 80% and 95% respectively. Japhet et al. [23] proposed a Fast-RCNN-based approach for ship detection [24]. Fast-RCNN is a Convolutional Neural Network (CNN) that is designed for faster training and testing paradigms. In this implementation, the authors used 400 images for training and testing and achieved a general accuracy of 87%. Huang et al. [26] introduced the BvSB-ADN algorithm that integrates active learning with deep learning, thus can achieve a high detection accuracy with fewer training samples. This method can select the best examples of training using deep network training. The network used in this study followed a Restricted Boltzmann Machine (RBM) structure [27]. Their method achieved an accuracy of 88.5% based on 200 training images.

The combination of Fast- RCNN and Region Proposal Network (RPN) is used to detect and select the best proposal of the image object. First Fast-RCNN calculates the feature map of the input image and then by using this feature map, RPN chooses some area as proposal objects. By combining the output of these two networks, the final proposal object are determined.

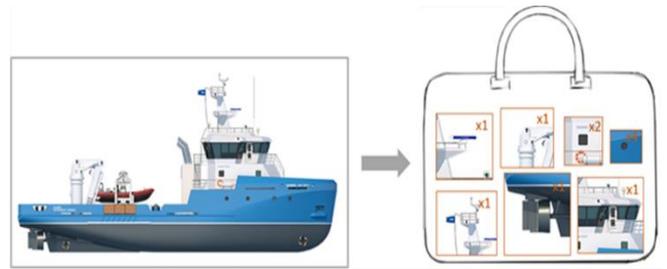

Fig. 3. A schematic view of the bag of words model

In recent years, many studies in the field of object detection have used a technique called the bag of words as a feature representation method for transferring the descriptive features in a way that they can be used with a suitable classifier such as SVM [28], [29] [30], [31], [32]. For example, to collapse the dimensionality of the features inside the bag of words, Lin et al. [31] presented a repetitive feature selection approach, such that these features are extracted from key points. Their method consists of two steps; the first of which was to define a number of key points represented as the reference in all of the images, and secondly, key points which their distances from these references are not greater than a predetermined limit, are selected from each image.

In this paper, we present an algorithm based on the Bag of Words concept for which the words are the features that are obtained by using pre-trained models of deep convolution networks. Two well-known and widely-used models in object recognition are applied in this paper including deep learning and SVM.

The rest of the paper is organized as follows: In section two, the description of our employed method is presented. Section 3 gives evaluations on the dataset and compares the results of our method with previous works and finally, section 4 provides the conclusion about the paper.

## II. METHOD

In this section, first, a brief description of the different modules used in the proposed method (e.g. bag of words and deep networks) are presented followed by a detailed explanation of the proposed method and finally evolution of our method is provided

### A. *The bag of words model*

This model provides an intuitive representation of image properties, such that the representation includes a more detailed look at the objects inside the image. The main characteristic of this model is looking at features as words, regardless of their locations, and only considering their frequency (i.e. the number of repetitions of a given word). Fig. 3 demonstrates a schematic overview of this model.

In this method, first, image features are extracted from the entire image region or segmented parts of the image and used as key information to describe the image. These features can be the ones retrieved based on the integration of gradient-based features [33] or other types of features based on color, texture, and form [35]. Then, feature vectors are first divided using a clustering method such as k-means into a predefined number of clusters. Afterward, the observation frequency of these feature vectors is considered as the new descriptive feature to be used by another classification network. In Fig. 4, the overview of this model is illustrated by a method based on the extraction of texture features.

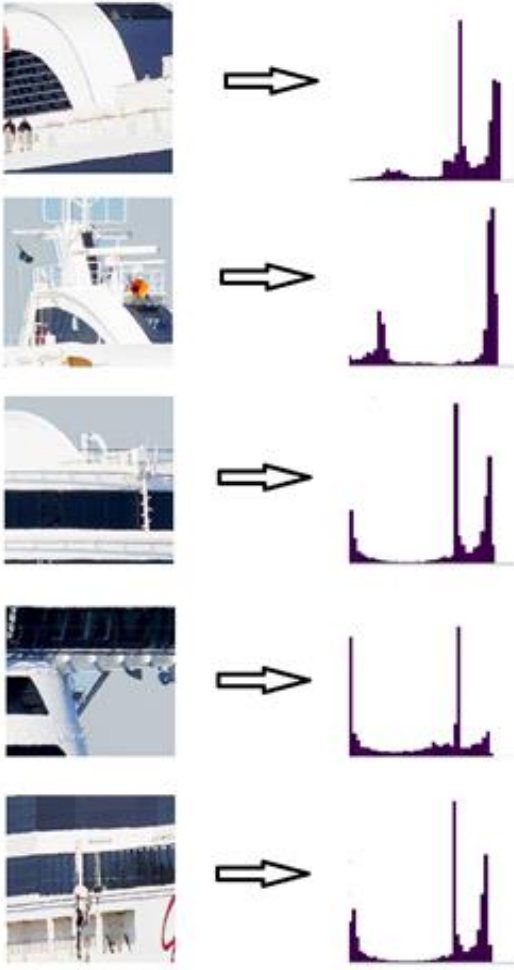

Fig. 4. BoW model for the method based on texture features

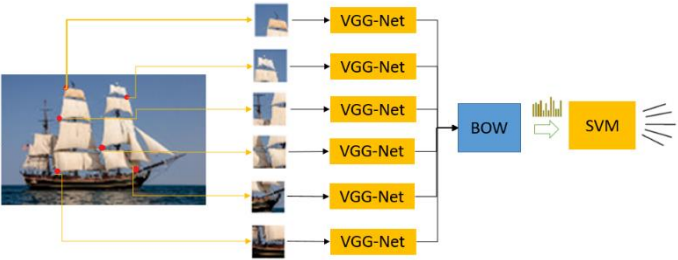

Fig. 5. General scheme of the proposed method. First, the key points in the input are detected by the SIFT method and then using an algorithm to choose a sub-sample of them. Then by using the VGG-Net we described an area around each point and make a BoW by them. Finally, by utilizing the BoW output and SVM classifier, the class of input ship is determined.

### B. Proposed method

The primary idea behind the proposed method is to extract appropriate key points from every image and to define a neighborhood around each point as input to a pre-trained Visual Geometry Group (VGG) network ([36]). The penultimate layer output is taken as a feature vector in this network. Finally, these vectors are used as feature vectors in the bag of words model. The general scheme of the algorithm is shown in Fig. 5.

In the next section, first, a general overview of the VGG network is presented; then, the details of the method are described in two stages of training and testing.

### C. VGG network

The VGG network provided by Chatfield et al. [44] is a 21-layers CNN, which the objective of learning and classifying a thousand categories of objects in the Image-Net [38] dataset. This network has eight convolution blocks so that the size of the filters starts from $7 \times 7$ and decreases to $3 \times 3$. The final layer, including 1000 neurons, is fully connected to the penultimate layer. This network, which has been rather successful in object detection on the Image-Net dataset [44] has three different versions, namely VGG-F (act), VGG-M (medium), and VGG-S (low).

In the proposed method, the output of the penultimate layer of the VGG-M-128 model is considered as a feature vector (descriptor). The reason behind choosing the VGG-M-128 model is that the number of features extracted from this model is the same as the number of features of the SIFT (128) method.

### D. Training Phase

First, key points were extracted from each image by using the SIFT method. Then, a $7 \times 7$ neighborhood is considered for every point followed by calculating the magnitude of the gradient for all the pixels within each neighborhood. Afterward, the key points were arranged based on the sum of the gradients and grouped them into a point set called $G_f$. Given the fact that we aimed to extract features from different areas of the image with the minimum overlap, we selected a certain number of important points to distribute the patches uniformly, such that their distance from each other was higher than a predetermined value, and put them in a set called $S_i$ which we updated each time from $S_0$ to $S_n$. Assuming that we represent the ordered set of points by $G_f$ and the first point in the set by $p_0$, we can define the $S_0$ set by the following:

$$S_0 = p_0 \in G_f \quad (1)$$

In the following, we update the $G_f$ and $S_{n+1}$ sets:

$$G_f = G_f - p_0, \quad (2)$$

$$\begin{cases} S_n = \{ p_0 \in G_f \mid N_{p_0} \geq |G_{n-1}| - minOver \} \\ N_{p_0} = \sum_{p_i \in S_{n-1}} \&(||p_i - p_0|| > DistTH) \end{cases}, \quad (3)$$

where, $N_{p_0}$ is the number of points in the set $S_n$ whose distance from the point $p_0$ is more than the DistTH threshold. $|S_n|$ and & represent the size of the $S_n$ set and the impulse function, respectively.

The condition $N_{p_0} \geq |S_n| - minOver$ states that if the number of points overlapping with the point under study, $p_0$, is less than a specified threshold ($minOver$ points), then this point is chosen as the new member of the final set. Evidently, if we set the $minOver$ value to zero, it means that we want all the points in the final chosen set to have a distance more than

DistTH; i.e., in this case, the maximum dispersion of points is considered. Each time, the sets $S_f$ and $S_{n+1}$ are updated using formulas (2) and (3) to meet the following condition:

$$|S_{n+1}| \geq TopN \quad (4),$$

Where $TopN$ is the number of important points from which we will calculate deep features around them. The points may be too close together or the number of key points may be low due to the uniformity of most of the image areas if all the points are processed. But the number of final points, designated as FoundN is less than $TopN$. So, by showing the number of remaining points with $RemainN$, These new points are selected from the first points of the final set, with the difference that the size of the neighborhood around these points is larger than before. The reason for using this method is that areas around the first key points contain more details. Therefore, by considering these areas larger, their features are extracted from a broader perspective.

Subsequently, after choosing all the points for every image, we extracted the deep features using the VGG network and added them to the new set called: Fea set. This set contains feature vectors obtained for all key points of all image sets.

Next, the Fea set is divided into N clusters by using the k-means method, so that the centers of these clusters are considered as a dictionary. Now, the BoW histogram is obtained by taking this dictionary as a basis. After calculating all histograms for each image of the dataset, and having its class label, selected from five different classes of ships, we can train the SVM classifier with the RBF kernel.

### III. 3. EXPERIMENTS AND RESULTS

#### A. Dataset

For this study, we used the dataset provided in [1]. We also tried to make the dataset more complete by referring to the Ship Spotting website, so that the number of images in this dataset increased from 1144 to 1400 color images. The dataset consists of five classes of ship types: container, gas, passenger, oil, and sailboats. Out of the dataset images, 1000 are used for the training phase and the remaining images are considered for evaluation. In Fig. 6, several images of each of the mentioned categories are shown. It worth noting that these images are used without any manipulation or preprocessing such as background segmentation in the proposed method.

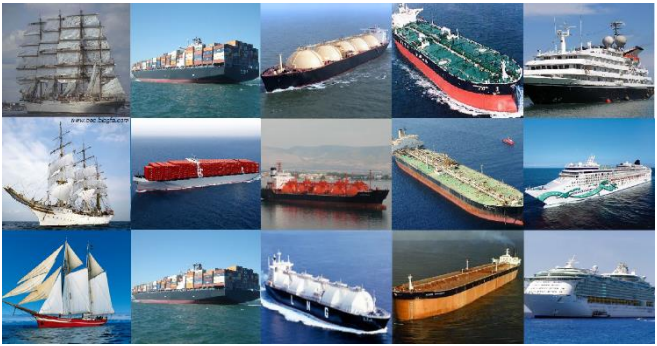

Fig. 6. Images of five ship categories (left to right): passenger, oil, gas, container, sailboats.

#### B. Evaluation criteria and results

To evaluate the results, the error evaluation criterion is defined as follows:

$$Er = \frac{1}{Total}\sum_{i=1}^{Total}\delta(t_i,c_i) \quad (5)$$

Where Total is the total number of images of training/testing sets and the function δ is the error evaluation function. So, if the output label ($c_i$) and the actual output ($c_i$) have the same value, δ has a value of 1; otherwise, it will be zero.

Similar to this study, two papers [1] and [26] used the same dataset and reported the highest output accuracy. It should be noted that the difference between the proposed method and [1] is that in SIFT, descriptors of this method are used as a feature vector, while in the proposed method, only the points obtained by the SIFT method are used. So, the proposed method is completely different. After clustering and selecting the most suitable points, a neighborhood is defined around each of the final points, and the deep features are extracted from each of these neighborhoods. Another difference is that, in the method [1], features are extracted from gray-level images, but in the proposed method, features are obtained from colored patches of RGB color space.

Also, the difference between the proposed method and the method of [26] is that they used the entire image as input to the Deep Belief Network called DBN, and also used the network itself as the classifier, while in the proposed method, the VGG network is used only to extract the feature (values obtained in the penultimate layer). Then, the support vector machine is applied for classification.

Empirically, the algorithm values for DistTH ،minOver ، TopN and Nbins should set to 15, 2, 100 ,and 50, respectively. To select these values, first, we considered the ones that seemed reasonable (10, 2, 150, and 100 in the order indicated above). Then, we evaluated training error and the testing error per increase or decrease of values of these parameters. Finally, we selected the best values. In the table (2), the proposed method error is shown for some parameter values.

It can be observed from Table 1 that the value of the DistTH parameter has the greatest effect on the accuracy of the proposed method because this parameter determines the dispersion rate and the distance of the selected area. In fact, an appropriate value for this parameter leads to the extraction of the features from the independent areas with the least overlapping.

In Table 2, the accuracy of the methods of [1], [26] and the proposed method is shown. The error for training and testing classes are shown as $Er_{train}$ and $Er_{test}$ The proposed algorithm shows a better improvement compared to other methods.

TABLE I. ERROR EVALUATION FOR DIFFERENT PARAMETER VALUES OF THE PROPOSED METHOD. ***DistTH***, ***minOver***, ***TopN***, ***Nbins***, $Er_{train}$ AND $Er_{test}$ ARE DISTANCE THRESHOLD BETWEEN THE SELECTED POINTS, MINIMUM POINTS NUMBER THAT CAN HAVE OVERLAP WITH OTHERS, NUMBER OF IMPORTANT POINTS WE SHOULD SELECT, NUMBER OF DICTIONARY WORDS, ERROR OF TRAINING AND ERROR OF TESTING RESPECTIVELY.

| ***DistTH*** | ***minOver*** | ***TopN*** | ***Nbins*** | $Er_{train}$ | $Er_{test}$ |
|---|---|---|---|---|---|
| 10 | 2 | 150 | 100 | 7.3 | 13.1 |
| 5 | 2 | 150 | 100 | 10.3 | 15.7 |
| 15 | 2 | 150 | 100 | 4.8 | 12.2 |
| 20 | 2 | 150 | 100 | 5.7 | 12.6 |
| 15 | 3 | 150 | 100 | 5.4 | 12.9 |
| 15 | 2 | 120 | 100 | 3.7 | 11.7 |
| 15 | 2 | 100 | 100 | 3.4 | 11.5 |
| 15 | 2 | 100 | 70 | 3.3 | 10.4 |
| 15 | 2 | 100 | 50 | 3.2 | 8.2 |

TABLE II. EVALUATION ERROR FOR THE METHODS OF [1], [26] AND THE PROPOSED METHOD

| Method | $Er_{train}$ | $Er_{test}$ |
|---|---|---|
| Method [1] | 8.4 | 15.2 |
| Method [26] | 7.6 | 11.4 |
| **Proposed method** | **3.2** | **8.2** |

## IV. DISCUSSION

In Fig. 7, some test images are illustrated which are classified correctly by the proposed method, but incorrectly with the other methods. After studying the results, we concluded that both methods of [1] and [26] face challenges in the classification of images that include other content than the target objects; meanwhile, the proposed method is affected less by this issue. The reason for the superiority of the proposed method compared to other methods is threefold.

First, unlike the method of [1], the points are not closely placed and the selected points are fairly uniformly distributed in the whole image, so all areas of the ships are equally considered for the extraction of the feature, and thus the BOW of each specific area has specific words.

Second, in the proposed method, the features that are chosen have a higher level of information than the SIFT features, therefore, they pay more attention to the color, texture and details.

Third, by using the SVM classifier instead of just using the last fully-connected layer of the deep network as the classifier, we achieved more nonlinearity in our method. This has leveraged higher accuracy in the complex object recognition like sub-category classifications.

Indeed, the features extracted by the VGG network have the potential for more generalization of the classifier since the VGG has been optimized for the recognition of many classes of objects. On the other hand, unlike the method of [26], the proposed method does not extract features from the whole image, but it extracts features from the important areas around the key points. So, since the BoW method is based on the histogram, it pays more attention to the repetition of these important areas in images of the training set.

In Fig. 8, some test images are shown that are not classified by the proposed method properly. As future improvements, this method should solve challenges when there are visual similarities between the recognized category and the actual category. Also, the proposed method must be able to classify the images that have other contents in addition to the target objects, even with its less error in comparison with the other two methods.

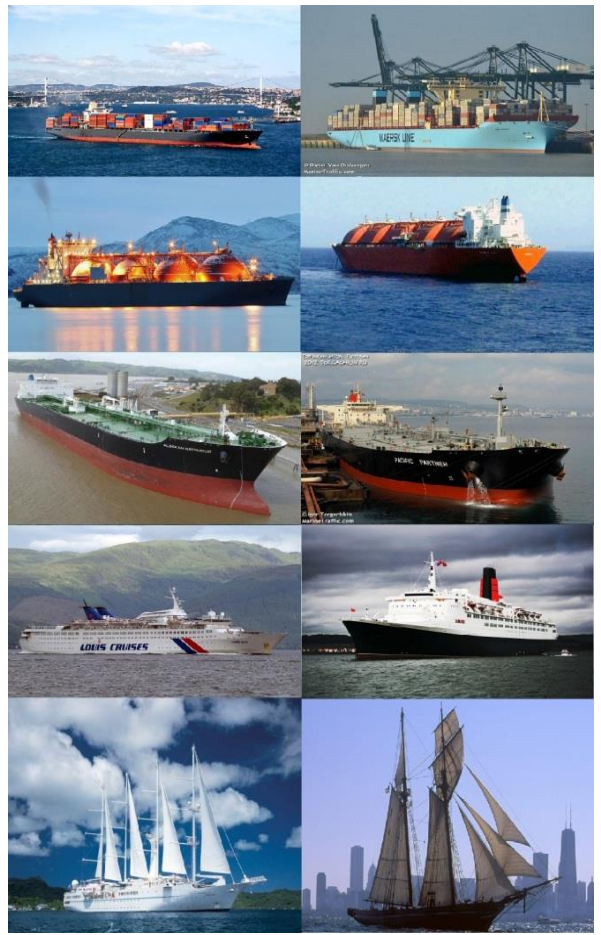

Fig. 7. Images from every class which are properly classified by the proposed method, but not properly classified by other methods

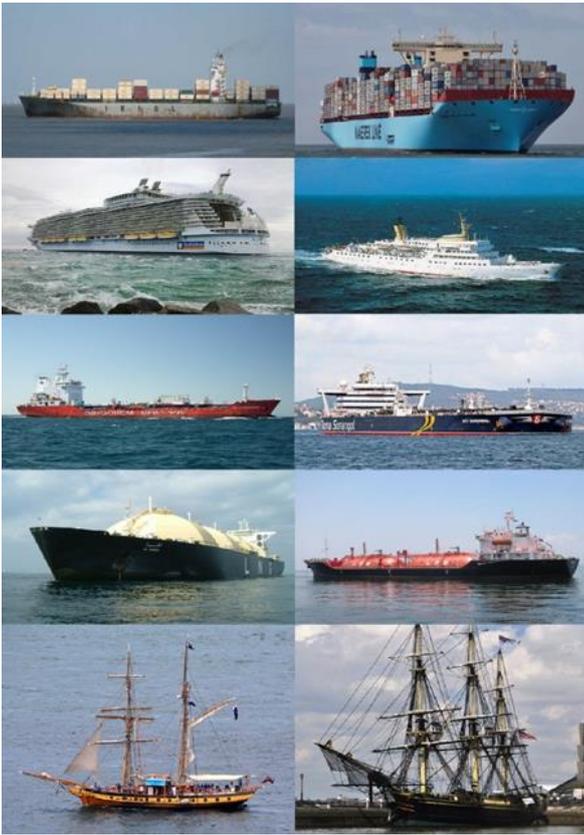

Fig. 8. Images of every category which is classified incorrectly by the proposed method

V. CONCLUSION

Recognition and classification of the ships according to their silhouette-profiles when crossing a marine passageway are among the areas that researchers have considered remarkably in terms of object recognition. In this paper, a method is presented based on the use of the BoW and deep features for image proposals. These image proposals were obtained using a greedy algorithm on key points that are selected using SIFT. The proposed method was tested on a dataset of 1400 images consisting of 5 different classes of ships. As observed in evaluations, the proposed algorithm shows a significant improvement compared to the existing methods. As future research, we aim to use a dimension reduction or feature selection method to reduce the dimensions of the final feature vector. Also, we will try to use a method to subtract the background of every non-ship object to increase accuracy. An idea for this approach is to use FCN networks [37].